\documentclass{llncs}
\usepackage[dvips]{graphics}
\usepackage{epsfig}
\usepackage{amsfonts}
\usepackage{amssymb}
\usepackage{amsmath}
\usepackage{color}
\usepackage{url}

\usepackage{algorithmic,algorithm}

\newcommand{\commentaire}[1]{ }

\newcommand{\Vois}[0]{{\cal N}}

\newcommand{\Real}{\mathop{\rm I\kern-.2emR}}
\newcommand{\Nat}{\mathop{\rm I\kern-.2emN}}
\renewcommand{\P}{\mathop{\rm I\kern-.2emP}}

\begin{document}

\mainmatter

\title{Pareto Local Optima of Multiobjective NK-Landscapes with Correlated Objectives}
\titlerunning{Pareto Local Optima of $\rho MNK$-Landscapes}

\author{S\'ebastien Verel\inst{1,3} \and Arnaud Liefooghe\inst{2,3} \and\\ Laetitia Jourdan\inst{3} \and Clarisse~Dhaenens\inst{2,3}}

\authorrunning{S. Verel, A. Liefooghe, L. Jourdan and C. Dhaenens}

\institute{
University of Nice Sophia Antipolis -- CNRS, France 
\and Universit\'e Lille 1, LIFL -- CNRS, France 
\and INRIA Lille-Nord Europe, France\\
\email{verel@i3s.unice.fr, arnaud.liefooghe@univ-lille1.fr, laetitia.jourdan@inria.fr, clarisse.dhaenens@lifl.fr}
}

\maketitle

\begin{abstract}
In this paper, we conduct a fitness landscape analysis for multiobjective combinatorial optimization,
based on the local optima of multiobjective $NK$-landscapes with objective correlation.
In single-objective optimization, it has become clear that
local optima have a strong impact on the performance of metaheuristics.
Here, we propose an extension to the multiobjective case, based on the Pareto dominance.
We study the co-influence of the problem dimension, 
the degree of non-linearity, the number of objectives
and the correlation degree between objective functions
on the number of Pareto local optima.
\end{abstract}

%======================================================================
\section{Motivations}
\label{sec:intro}

%A single term for : local search, metaheuristics, search approaches\dots ?

The aim of fitness landscape analysis is to understand the properties of a given combinatorial optimization problem
in order to design efficient search algorithms.
One of the main feature is related to the number of local optima,
to their distribution over the search space and to the shape of their basins of attraction.
For instance, in single-objective optimization, it has been shown that local optima tend to be clustered in a `central massif'
for numerous combinatorial problems, such as the family of  $NK$-landscapes~\cite{kauffman93}.
A lot of methods are designed to `escape' from such local optima.
However, very little is known in the frame of multiobjective combinatorial optimization (MoCO),
where one of the most challenging question relies on the identification of the set of Pareto optimal solutions.
A Pareto Local Optima (PLO) \cite{paquete2007a} is a solution that is not dominated by any of its neighbors.
The description of PLO is one of the first fundamental step towards the description of the structural properties of a MoCO problem.
Surprisingly,  up to now, 
there is a lack of study on the number and on the distribution of PLO in MoCO.

Like in single-objective optimization, 
the PLO-related properties clearly have a strong impact on the landscape of the problem, and then on the efficiency of search algorithms.
In particular, local search algorithms are designed in order to take them into account.
For instance, the family of Pareto Local Search (PLS)~\cite{paquete2007a} iteratively improves a set of solutions
with respect to a given neighborhood operator and to the Pareto dominance relation.
The aim of PLS, like a number of other search algorithms, is to find a set of mutually non-dominated PLO.
PLS has been proved to terminate on such a set, called a \emph{Pareto local optimum set}~\cite{paquete2007a}.
Notice that a Pareto optimal solution is a PLO, and that
the whole set of Pareto optimal solutions is a Pareto local optimum set. 
The behavior of multiobjective algorithms clearly depends on the properties related to the PLO.
First, a Pareto local optimum set is always a subset of the whole set of PLO.
Second, the dynamics of a PLS-like algorithm depends of the number of PLO found along the search process.
The probability to improve an approximation set that contains a majority of PLO should be smaller than
the probability to improve an approximation set with few PLO.

%Since the seminal work of Knowles and Corne \cite{knowles2002},
%a very small amount of literature related to fitness landscape exists for MoCO.
There exists a small amount of literature related to fitness landscape for MoCO.
Borges and Hansen \cite{borges1998} study the distribution of local optima, in terms of scalarized functions,
for the multiobjective traveling salesman problem (TSP).
Another analysis of neighborhood-related properties for biobjective TSP instances of different structures is given in \cite{paquete2009}.
Knowles and Corne \cite{knowles2002} lead a landscape analysis
on the multiobjective quadratic assignment problem with a rough objective correlation.
Next, the transposition of standard tools from fitness landscape analysis to MoCO are discussed by Garrett \cite{garrett2007}, 
and an experimental study is conducted with fitness distance correlation.
But this measure requires the true Pareto optimal set to be known. % (or at least a good approximation of it).
In another study, the landscape of a MoCO problem is regarded as a neutral landscape, 
and divided into different fronts with the same dominance rank \cite{Garrett09}.
In such a case, a small search space needs to be enumerated.
In previous works on multiobjective $NK$-landscapes by Aguirre and Tanaka~\cite{aguirre2007},
small enumerable fitness landscapes are studied according to the number of fronts, the number of solutions on each front,
the probability to pass from one front to another, and the hypervolume of the Pareto front. 
However, the study of fronts simply~allows to analyze small search spaces,
and from the point of view of dominance rank~only.

In this work, our attempt is to analyze the structure of large search space using the central notion of local optimum.
For the design of a local search algorithm for MoCO,  the following questions are under study in this paper:
($i$)~What is the number of PLO in the whole search space?
($ii$)~Is the number of PLO related to the number of Pareto optimal solutions?
In particular we want to study such properties according to the correlation degree between objective functions.
In order to study the problem structure, and in particular the PLO, 
we use the multiobjective $NK$-landscapes with objective correlation, \emph{$\rho MNK$-landscapes} for short,
recently proposed in~\cite{verel2011}.
The contributions of this work can be summarized as follows.
First, we show the co-influence of objective correlation, objective space dimension and epistasis on the number of PLO.
Next, we propose a method based on the length of a Pareto adaptive walk to estimate this number.
At last, we study the number of PLO for large-size instances.

The paper is organized as follows. 
Section \ref{sec:lsmoco} deals with MoCO and local search algorithms.
Section \ref{sec:rhomnk} is devoted to the definition of multiobjective $NK$-landscapes with objective correlation.
In Section \ref{sec:plo}, 
we study the number of PLO for enumerable instances and we propose a method to estimate it.
Moreover, we analyze the correlation between the number of PLO and Pareto optimal solutions.
In Section \ref{sec:large}, the co-influence of objective space dimension, objective correlation and epistasis is studied
for the PLO of large-size instances.
The last section concludes the paper.

%======================================================================
\section{Local Search for Multiobjective Combinatorial Optimization}
\label{sec:lsmoco}
%This section introduces definitions and notations for multiobjective combinatorial optimization (MoCO),
%discusses the design of metaheuristics for MoCO and presents single-objective and multiobjective $NK$-landscapes.

%--------------------------------------------------
\subsection{Multiobjective Combinatorial Optimization}
\label{sec:moco}
A multiobjective combinatorial optimization (MoCO) problem can be defined by
a set of $M \geq 2$ objective functions $(f_1, f_2,\dots, f_M)$,
and a (discrete) set $X$ of feasible solutions in the \emph{decision space}.
Let $Z  =  f(X) \subseteq \Real^M$ be the set of feasible outcome vectors in the  \emph{objective space}.  
In a maximization context, a solution $x^\prime \in X$ is dominated by a solution $x \in X$, denoted by $x^\prime \prec x$,
iff $\forall i \in \{1,2,\dots,M\}$, $f_i(x^\prime) \leq f_i(x)$ and $\exists j \in \{1,2,\dots,M\} $ such that $f_j(x^\prime) < f_j(x)$.
A solution $x \in X$ is said to be \emph{Pareto optimal} (or \emph{efficient}, \emph{non-dominated}), 
if there does not exist any other solution $x^\prime \in X$ such that $x^\prime$ dominates $x$.
The set of all Pareto optimal solutions is called the \emph{Pareto optimal set} (or the \emph{efficient set}), denoted by $X_E$,
and its mapping in the objective space is called the \emph{Pareto front}.
A possible approach in MoCO is to identify the minimal complete Pareto optimal set,
{\itshape i.e.}~one solution mapping to each point of the Pareto front.
However, the overall goal is often to identify a good Pareto set approximation.
To this end, metaheuristics in general, and evolutionary algorithms in particular, have received a growing interest since the late eighties.
Multiobjective metaheuristics still constitute an active research area~\cite{coello2010}.
%\cite{coello2007,coello2010}.

%--------------------------------------------------
\subsection{Local Search}
\label{sec:ls}
% ----- neighborhood
A \emph{neighborhood structure} is a function $\mathcal{N} :
X \rightarrow 2^X$ that assigns a set of solutions $\mathcal{N}(x) \subset X$ to any solution $x \in X$.
The set $\mathcal{N}(x)$ is called the \emph{neighborhood} of $x$, and a solution $x' \in \mathcal{N}(x)$ is called a \emph{neighbor} of $x$.
% ----- local optima
In single-objective combinatorial optimization, a fitness landscape can be defined by the triplet $(X, \mathcal{N}, h)$,
where $h : X \longrightarrow \Real$ represents the fitness function,
that can be pictured as the \textit{height} of the corresponding solutions.
Each peak of the landscape corresponds to a local optimum.
In a single-objective maximization context,
a \emph{local optimum} is a solution $x^{\star}$ such that $\forall x \in \mathcal{N}(x^{\star})$, $f(x) \leq f(x^{\star})$.
The ability of local search algorithms  has been shown to be related to the number of local optima for the problem under study, 
and to their distribution over the landscapes \cite{merz2004}.

% ----- PLO = PARETO LOCAL OPTIMA
In MoCO,  given that Pareto optimal solutions are to be found,
the notion of local optimum has to be defined in terms of Pareto optimality.
Let us define the concepts of Pareto local optimum and of Pareto local optimum set.
For more details, refer to \cite{paquete2007a}.
A solution $x \in X$ is a \emph{Pareto local optimum} (PLO) with respect to a neighborhood structure~$\mathcal{N}$
if there does not exist any neighboring solution $x' \in \mathcal{N}(x)$ such that $x \prec x^\prime$.
% ----- PLO set
A \emph{Pareto local optimum set} $X_{PLO} \in X$ with respect to a neighborhood structure~$\mathcal{N}$
is a set of mutually non-dominated solutions such that
$\forall x \in X_{PLO}$, there does not exist any solution $x^{\prime} \in \mathcal{N}(X_{PLO})$ such that $x \prec x^\prime$.
In other words, a locally Pareto optimal set cannot be improved, in terms of Pareto optimality, by adding solutions from its neighborhood.
%In other words, a locally Pareto optimal set does not have any neighboring solutions such that ... ?

% ----- algo
Recently, local search algorithms have been successfully applied to MoCO problems.
%Some properties of the landscape seem to allow the search space to be explored in an effective way.
Such methods seem to take advantage of some properties of the landscape in order to explore the search space in an effective way.
Two main classes of local search for MoCO can be distinguished.
The first ones, known as \emph{scalar approaches}, are based on multiple scalarized aggregations of the objective functions.
The second ones, known as \emph{Pareto-based approaches}, directly or indirectly focus the search on the Pareto dominance relation
(or a slight modification of~it).
One of them is the Pareto Local Search (PLS) \cite{paquete2007a}.
It combines the use of a neighborhood structure
with the management of an archive (or population) of mutually non-dominated solutions found so far.
The basic idea is to iteratively improve this archive by exploring the neighborhood of its own content until no~further improvement is possible,
{\itshape i.e.} the archive falls in a Pareto local optimum set~\cite{paquete2007a}.

%======================================================================
\section{$\rho MNK$-Landscapes: Multiobjective $NK$-Landscapes with Objective Correlation}
\label{sec:rhomnk}
In single-objective optimization,
the family of $NK$-landscapes constitutes an interesting model to study the influence of non-linearity on the number of local optima.
In this section, we present the $\rho MNK$-landscapes proposed in \cite{verel2011}.
They are based on the $MNK$-landscapes \cite{aguirre2007}.
In this multiobjective model, the correlation between objective functions can be precisely tuned by a correlation parameter value.

%--------------------------------------------------
\subsection{$NK$- and $MNK$-Landscapes}
The family of $NK$-landscapes \cite{kauffman93} is a problem-independent model used for constructing multimodal landscapes.
$N$ refers to the number of (binary) genes in the genotype ({\itshape i.e.} the string length)
and $K$ to the number of genes that influence a particular gene from the string (the epistatic interactions).
By increasing the value of $K$ from 0 to $(N-1)$, $NK$-landscapes can be gradually tuned from smooth to rugged.
The fitness function (to be maximized) of a $NK$-landscape $f_{NK}: \lbrace 0, 1 \rbrace^{N} \rightarrow [0,1)$ is defined on binary strings of size $N$.
An `atom' with fixed epistasis level is represented by a fitness component $f_i: \lbrace 0, 1 \rbrace^{K+1} \rightarrow [0,1)$ associated to each bit $i \in N$.
Its value depends on the allele at bit $i$ and also on the alleles at $K$ other epistatic positions ($K$ must fall between $0$ and $N - 1$).
The fitness $f_{NK}(x)$ of a solution $x \in \lbrace 0, 1 \rbrace^{N}$ corresponds to the mean value of its $N$ fitness components $f_i$: \label{defNK}
$ f_{NK}(x) = \frac{1}{N} \sum_{i=1}^{N} f_i(x_i, x_{i_1}, \ldots, x_{i_K})$,
where $\lbrace i_1, \ldots, i_{K} \rbrace \subset \lbrace 1, \ldots, i-1, i+1, \ldots, N \rbrace$.
In this work, we set the $K$ bits randomly on the bit string of size $N$.
Each fitness component $f_i$ is specified by extension,
\textit{i.e.} a number $y^i_{x_i, x_{i_1}, \ldots, x_{i_K}}$ from $[0, 1)$ is associated with each element
$(x_i, x_{i_1}, \ldots, x_{i_K})$ from $\lbrace 0, 1 \rbrace^{K+1}$.
Those numbers are uniformly distributed in the range $[0, 1)$. 

More recently, a multiobjective variant of $NK$-landscapes (namely $MNK$- landscapes) \cite{aguirre2007}
has been defined with a set of $M$ fitness functions:
$$\forall m \in [1,M],\ f_{NK_{m}}(x) = \frac{1}{N} \sum_{i=1}^{N} f_{m, i}(x_i, x_{i_{m, 1}}, \ldots, x_{i_{m, K_{m}}})$$
The numbers of epistasis links $K_{m}$ can theoretically be different for each fitness function. 
But in practice, the same epistasis degree $K_{m}=K$ for all $m \in [1,M]$ is used.
Each fitness component $f_{m, i}$ is specified by extension
with the numbers $y^{m,i}_{x_i, x_{i_{m, 1}}, \ldots, x_{i_{m, K_m}}}$.
In the original $MNK$-landscapes \cite{aguirre2007}, these numbers are randomly and independently drawn from $[0, 1)$.
As a consequence, it is very unlikely that two different solutions map to the same point in the objective space.

%--------------------------------------------------
\subsection{$\rho MNK$-Landscapes}

In \cite{verel2011},  $CMNK$-landscapes have been proposed.
The epistasis structure is identical for all the objectives:
$\forall m \in [1,M]$,  $K_{m}=K$
and
$\forall m \in [1,M]$, $\forall j \in [1,K]$, $i_{m, j} = i_{j}$.
The fitness components are not defined independently.
The numbers $(y^{1,i}_{x_i, x_{i_1}, \ldots, x_{i_{K}}}, \ldots, y^{M,i}_{x_i, x_{i_1}, \ldots, x_{i_{K}}})$ follow a multivariate uniform law of dimension~$M$,
defined by a correlation matrix $C$.
Thus, the $y$'s follow a multidimensional law with uniform marginals 
and the correlations between $y^{m, i}_{\ldots}$s are defined by the matrix $C$.
So, the four parameters of the family of $CMNK$-landscapes are 
($i$)~the number of objective functions $M$,
($ii$)~the length of the bit string $N$,
($iii$)~the number of epistatic links $K$, 
and ($iv$)~the correlation matrix $C$.

In the $\rho MNK$-landscapes,
a matrix $C_{\rho}=(c_{np})$ is considered, with the same correlation between all the objectives:
$c_{nn} = 1$ for all $n$, and $c_{np} = \rho$ for all $n \not= p$.
However, it is not possible to have the matrix $C_{\rho}$ for all $\rho$ between $[-1,1]$:
$\rho$ must be greater than $\frac{-1}{M-1}$, see \cite{verel2011}.
To generate random variables with uniform marginals and a specified correlation matrix $C$, we follow the work of
Hotelling and Pabst \cite{hotelling1936}.
The construction of $CMNK$-landscapes defines correlation between the $y$'s but not directly between the objectives. 
In \cite{verel2011}, it is proven by algebra that the correlation between objectives is tuned by the matrix $C$:
$E(cor(f_{n}, f_{p})) = c_{np}$.
In $\rho MNK$-landscape, the parameter $\rho$ allows to tune very precisely the correlation between all pairs of objectives.

%*************************************************************************************
\section{Study of Pareto Local Optima}
\label{sec:plo}
%*************************************************************************************

In this section, we first study the number of Pareto local optima (PLO) according 
to the objective correlation, the number of objectives and the epistasis of  $\rho MNK$-landscapes.
Then, we analyze its relation with the size of the Pareto optimal set.
At last, we propose an adaptive walk that is able to estimate the number of PLO very precisely.
We conduct an empirical study for $N=18$ so that we can enumerate all the PLO exhaustively.
In order to minimize the influence of the random creation of landscapes, 
we considered $30$ different and independent instances for each parameter combinations: $\rho$, $M$, and $K$.
The measures reported are the average over these 30 landscapes. 
The parameters under investigation in this study are given in Table~\ref{tab:param}.
% table
\begin{table}[t]
\caption{Parameters used in the paper for the experimental analysis.}
\begin{center}
\begin{tabular}{c|lll}
Parameter & Values \\
\hline
$N$ & $\{18\} \, $ (Section~\ref{sec:plo}) &, & $\{18, 32, 64, 128\} \,$ (Section \ref{sec:large}) \\
$M$ & $\{2, 3, 5 \}$ \\
$K$ & $\{2, 4, 6, 8, 10 \}$\\
$\rho$ &  \multicolumn{3}{l}{$\{ -0.9, -0.7, -0.4, -0.2, 0.0, 0.2, 0.4, 0.7, 0.9 \}$  such that $\rho \geq \frac{-1}{M-1}$} \\
\end{tabular}
\end{center}
\label{tab:param}
\end{table}

%======================================================
\subsection{Number of Pareto Local Optima}

\begin{figure} [t]
\begin{center}
\begin{tabular}{cc}
\includegraphics[width=0.45\textwidth]{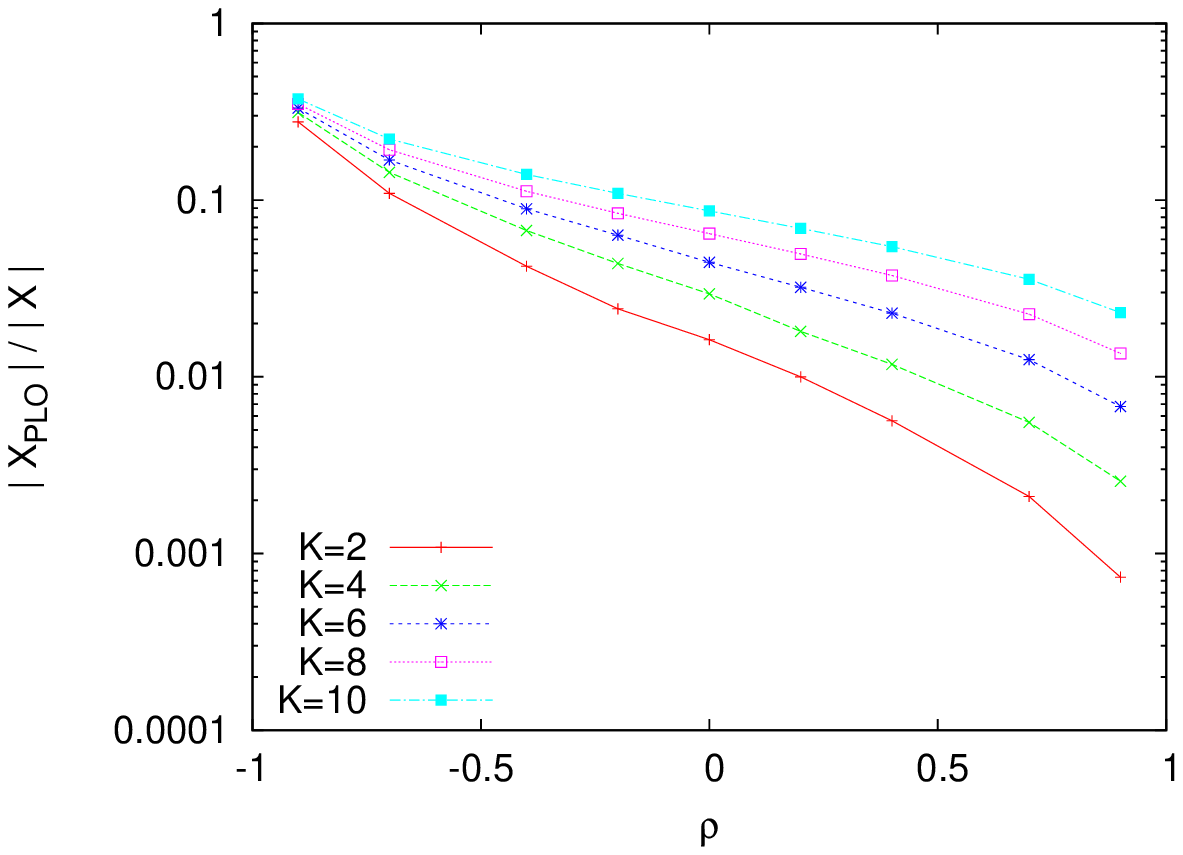} & \includegraphics[width=0.45\textwidth]{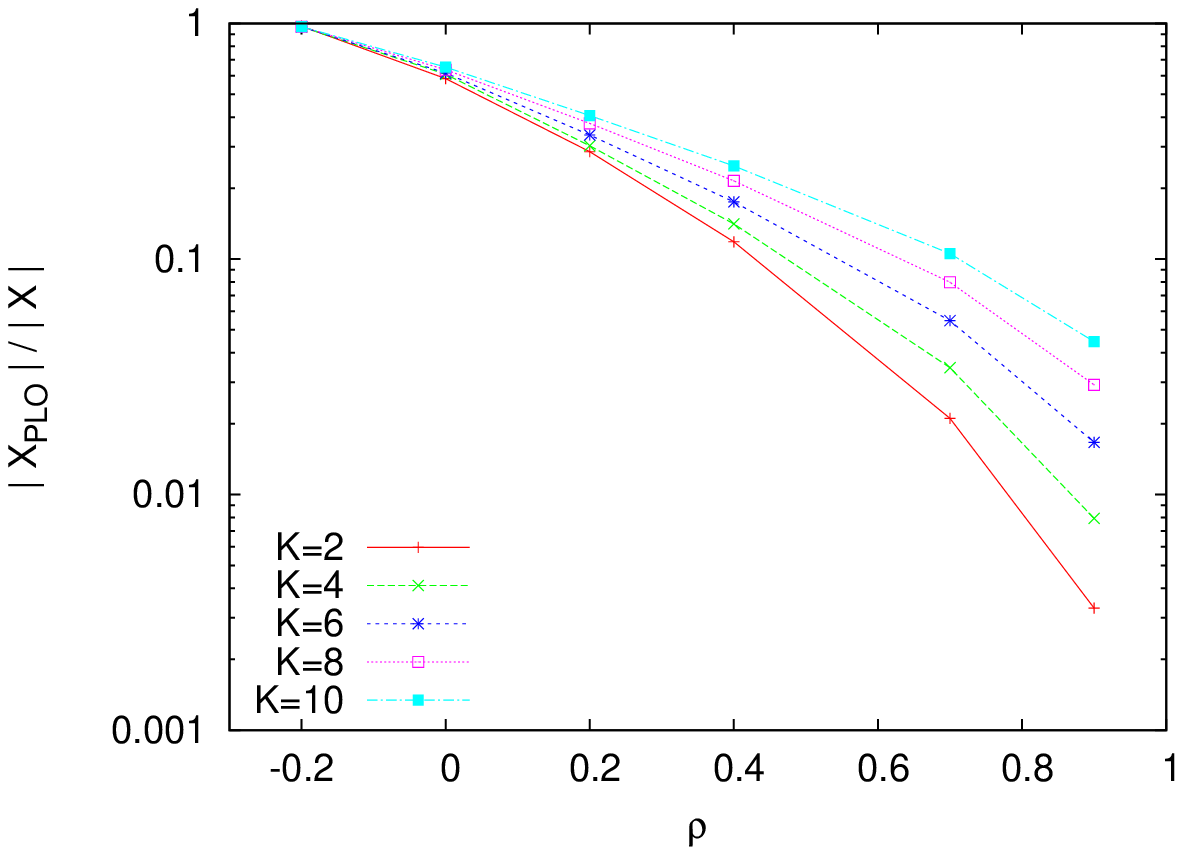} \\
%\multicolumn{2}{c}{\includegraphics[width=0.45\textwidth]{nbPLO_5_log.eps}} \\
\includegraphics[width=0.45\textwidth]{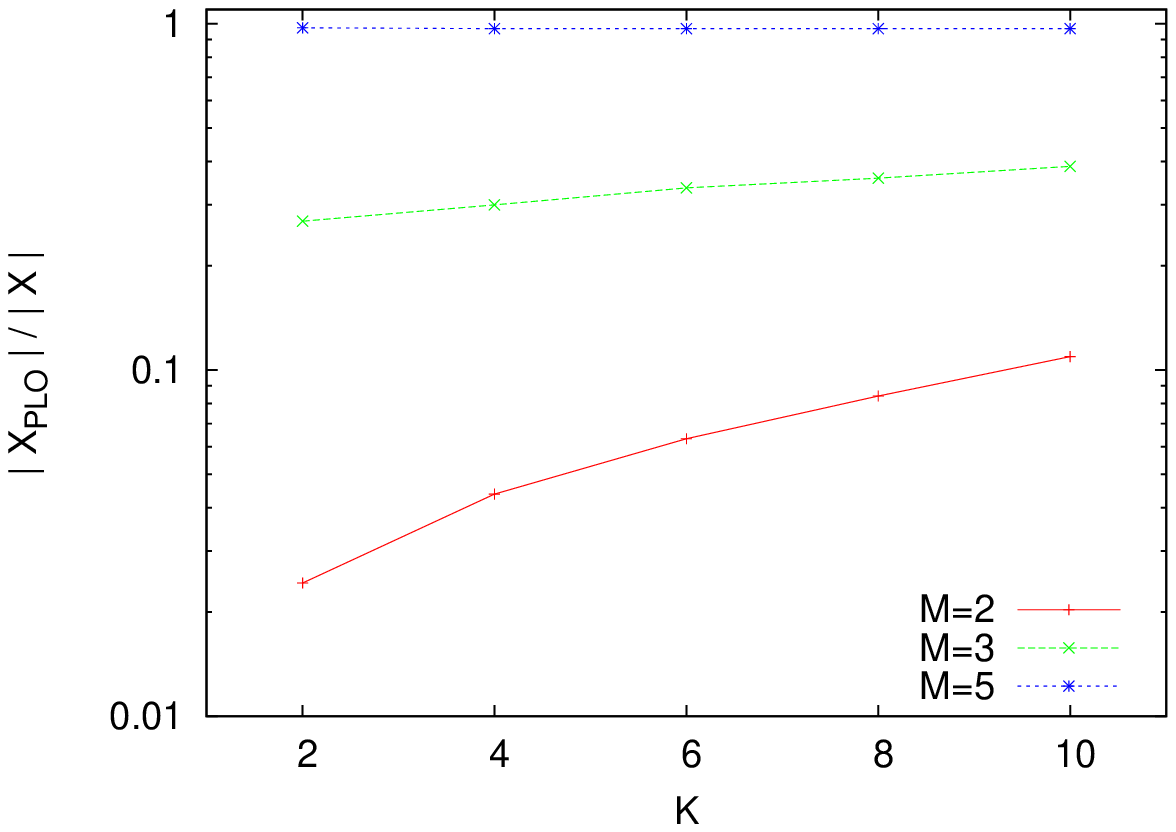} & \includegraphics[width=0.45\textwidth]{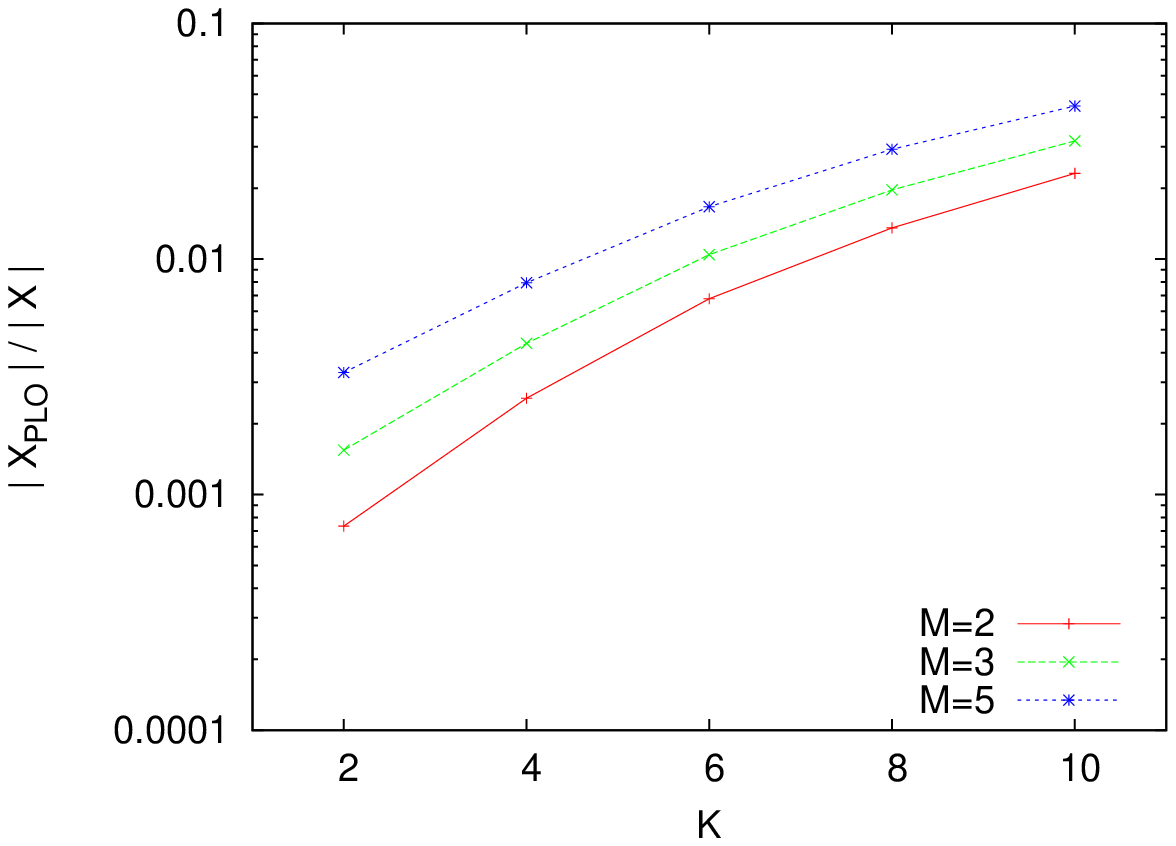} \\
\end{tabular}
\caption{Average number of PLO to the size of the search space~($|X| = 2^{18}$)
according to parameter $\rho$ (top left $M=2$, right $M=5$),
and to parameter $K$ (bottom left $\rho=-0.2$, right $\rho=0.9$).
The problem size is $N=18$.
\label{fig:nbPlo}}
\end{center}
\end{figure}

%First, let us remind that a Pareto Local Optimum (PLO) is a solution with no neighbor that dominates it (see Section \ref{sec:ls}).
Fig. \ref{fig:nbPlo} shows the average number of PLO to the size of the search space ($| X | = 2^{18}$)
for different $\rho$MNK-landscapes parameter settings.
As the well-known result from single-objective $NK$-landscapes \cite{kauffman93},
the number of PLO increases with the epistasis degree.
For instance, with an objective space dimension $M=2$ and an objective correlation $\rho = 0.9$,
the average number of PLO increases more than $30$ times: from $192$ for $K=2$ to $6048$ for $K=10$.
However, the range of PLO is larger with respect to objective correlation. 
For the same epistatic degree and number of objectives, 
the number of PLO decreases exponentially (Fig. \ref{fig:nbPlo}, top).
Indeed, for an objective space dimension $M=2$ and an epistasis degree $K=4$,
the average number of PLO decreases more than $120$ times:
from $82,093$ for negative correlation ($\rho=-0.9$) to $672$ for positive correlation ($\rho=0.9$).

This result can be interpreted as follows.
Let us consider an arbitrary solution~$x$, and two different objective functions $f_i$ and $f_j$.
When the objective correlation is high, there is a high probability that $f_i(x)$ is close to $f_j(x)$.
In the same way,
the fitness values $f_i(x^\prime)$ and $f_j(x^\prime)$ of a given neighbor $x^\prime \in \Vois(x)$ are probably close.
So, for a given solution $x$ such that it exists a neighbor $x^\prime \in \Vois(x)$ with a better $f_i$-value,
the probability is high that $f_j(x^\prime)$ is better than $f_j(x)$.
More formally, the probability $\P(f_j(x^\prime)>f_j(x) ~|~ f_i(x^\prime)>f_i(x))$, with $x^\prime \in \Vois(x)$,
increases with the objective correlation. 
Then, a solution $x$ has a higher probability of being dominated when the objective correlation is high.
Under this hypothesis,
the probability that a solution dominates all its neighbors 
decreases with the number of objectives.
Fig.~\ref{fig:nbPlo} (bottom) corroborates this hypothesis.
When the objective correlation is negative ($\rho=-0.2$),
the number of PLO changes in an order of magnitude from $M=2$ to $M=3$, and from $M=3$ to $M=5$.
This range is smaller when the correlation is positive.
When the number of objective is large and the objective correlation is negative, almost all solutions are PLO.

Assuming that the difficulty for Pareto-based search approaches gets higher when the number of PLO is large,
the difficulty of  $\rho MNK$-landscapes increases when:
($i$) the epistasis increases,
($ii$) the number of objective functions increases,
($iii$) the objective correlation is negative, and its absolute value increases.
Section~\ref{sec:large} will precise the relative difficulty related to those parameters for large-size problem instances.

% figure

%======================================================
\subsection{Estimating the Cardinality of the Pareto Optimal Set?}
\label{sec:est}

When the number of Pareto optimal solutions is too large, 
it becomes impossible to enumerate them all.
A metaheuristic should then manipulate a limited-size solution set during the search.
In this case, we have to design specific strategies to limit the size of the approximation set~\cite{knowles2004}.
Hence, the cardinality of the Pareto optimal set also plays a major role in the design of multiobjective metaheuristics.

\begin{figure}[t]
\begin{center}
\includegraphics[width=0.6\textwidth]{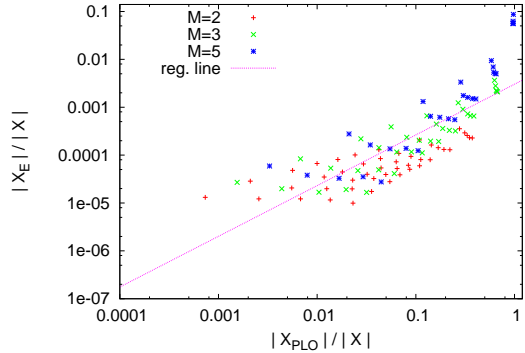}
\caption{Scatter plot of the average size of the Pareto optimal set (to the size of the search)
\textit{vs.} the average number of PLO (to the size of the search) for the $110$ possible combinations of parameters.
The problem size is $N=18$.
The correlation coefficient is~$0.82$. Notice the log-scales.\label{fig:nbFPF_nbPLO}}
\end{center}
\end{figure}

In order to design such an approach, 
it would be convenient to approximate the size of the Pareto optimal set from the number of PLO.
Fig.~\ref{fig:nbFPF_nbPLO} shows the scatter plot of the average size of the Pareto optimal set
\textit{vs.} the average number of PLO in log-scales.
Points are scattered over the regression line with the Spearson correlation coefficient of $0.82$, %$r = 0.819988$
and the regression line equation is $\log(y) = a \log(x) + b$ with $a=1.059$ and $b=-6.536$. % $a=1.05901$ and $b=-6.5354$.
For such a log-log scale, the correlation is low.
It is only possible to estimate the cardinality of the Pareto optimal set from the number of PLO with a factor $10$.
Nevertheless, 
the number of Pareto optimal solutions clearly increases when the number of PLO increases.

% figure

%======================================================
\subsection{Adaptive Walk}

In single-objective optimization, the length of adaptive walks, performed with a hill-climber,
allows to estimate the diameter of the local optima basins of attraction.
Then, the number of local optima can be estimated when the whole search space cannot be enumerated exhaustively.
In this section, we define a multiobjective hill-climber,
and we show that the length of the corresponding adaptive walk is correlated to the number of PLO.
We define a very basic single solution-based \textit{Pareto Hill-Climbing} (PHC) for multiobjective optimization.
A pseudo-code is given in Algorithm \ref{algo:phc}.
At each iteration of the PHC algorithm, the current solution is replaced by one random neighbor solution which dominates it.
So, the PHC stops on a PLO.
The number of iterations, or steps, of the PHC algorithm is the length of the Pareto adaptive walk.

% algo
\begin{algorithm}[b]
\caption{Pareto Hill-Climbing (PHC)}
\label{algo:phc}
\begin{algorithmic}
\STATE start with a random solution $x \in X$ 
\STATE step $\leftarrow$ $0$
\WHILE{$x$ is not a Pareto Local optimum}
    \STATE randomly choose  $x^{'}$ from $ \{ y \in \Vois(x) | x \prec y \}$
    \STATE $x \leftarrow x^{'}$
%    \STATE choose $s^{'} \in \Vois(s)$ such that $s^{'}$ dominate $s$
    \STATE step $\leftarrow$ step $+ 1$
\ENDWHILE
\end{algorithmic}
\end{algorithm}

\begin{figure}[t]
\begin{center}
\begin{tabular}{cc}
\includegraphics[width=0.45\textwidth]{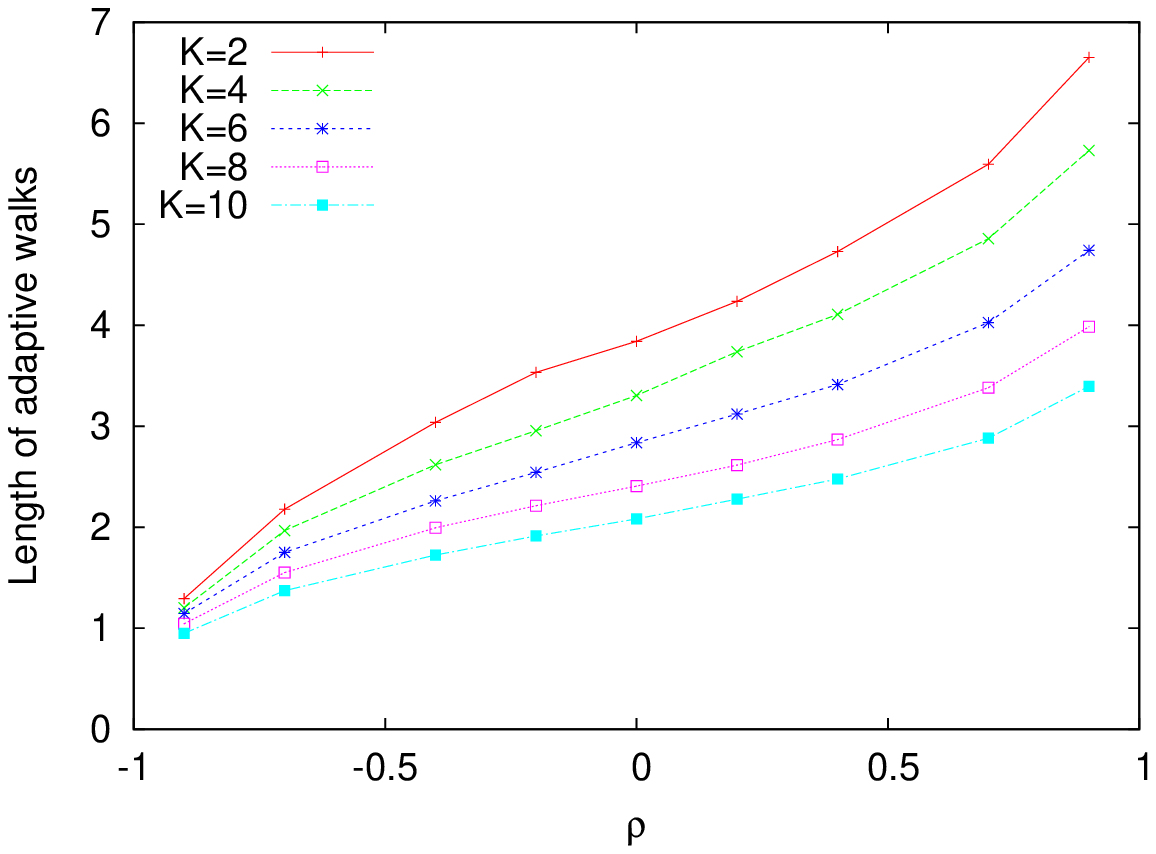} & \includegraphics[width=0.45\textwidth]{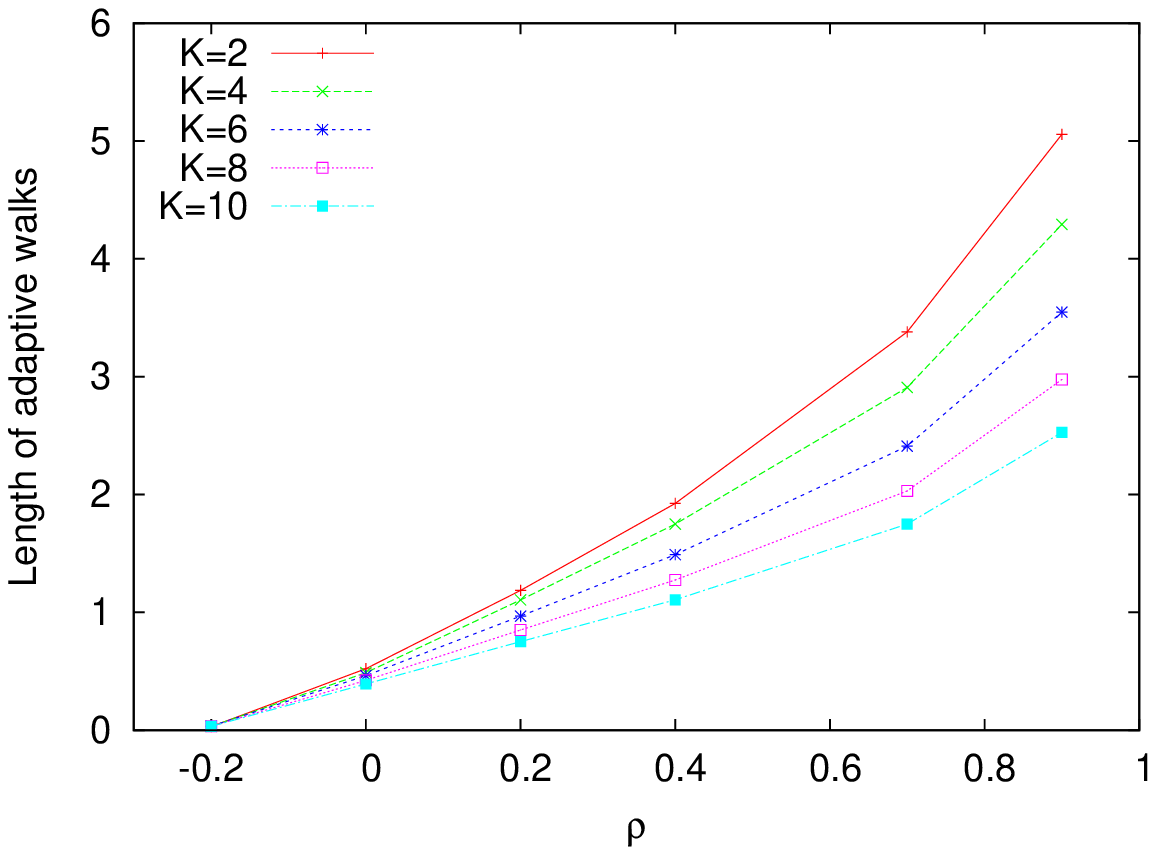} \\
\includegraphics[width=0.45\textwidth]{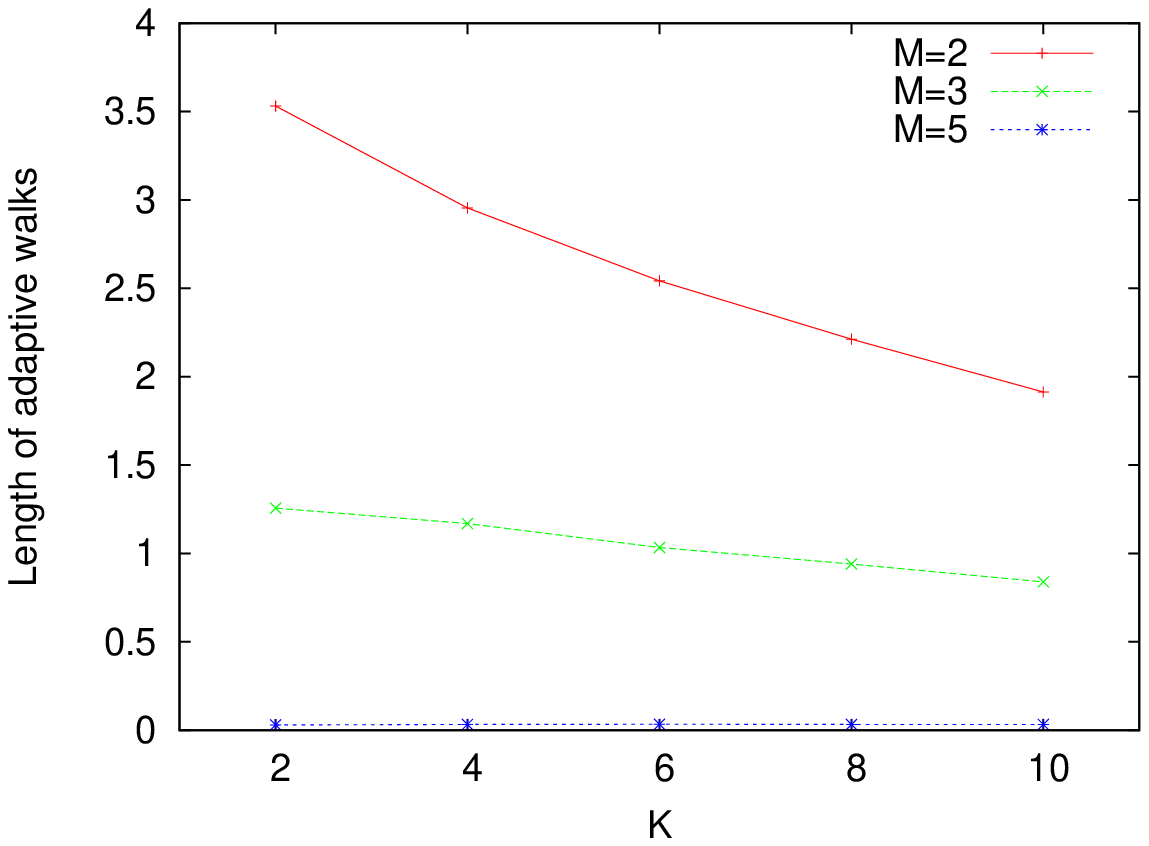} & \includegraphics[width=0.45\textwidth]{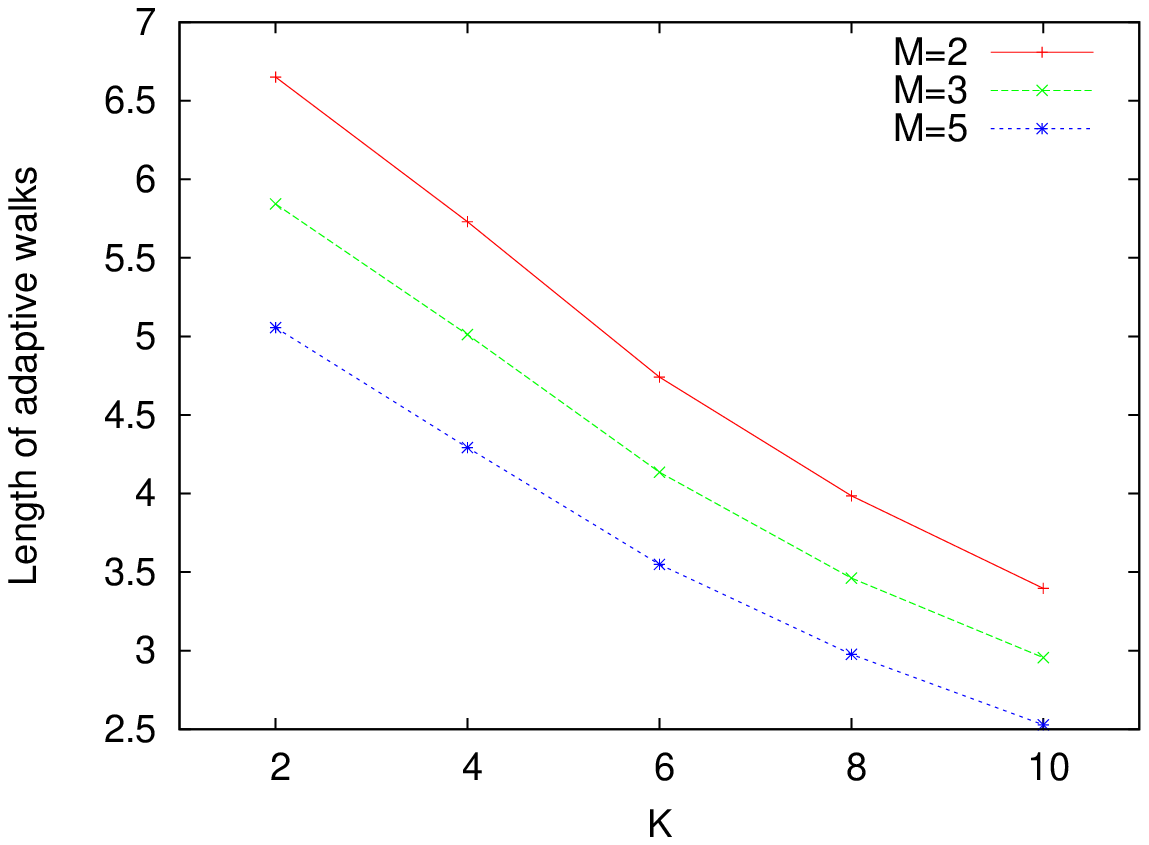} \\
\end{tabular}
\caption{Average length of the Pareto adaptive walk
according to parameter $\rho$ (top left $M=2$, right $M=5$),
and according to parameter $K$ (bottom left $\rho=-0.2$, right $\rho=0.9$).
The problem size is $N=18$.
\label{fig:lengthPHC}}
\end{center}
\end{figure}

\begin{figure}[t]
\begin{center}
\includegraphics[width=0.6\textwidth]{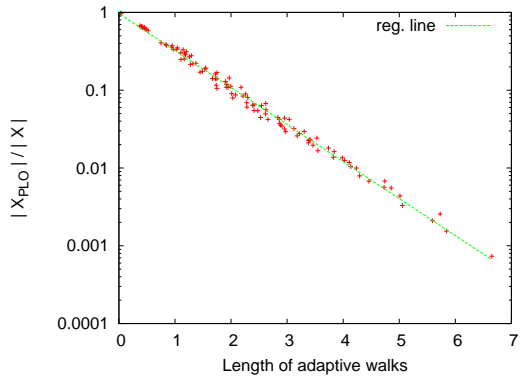} 
\caption{Scatter plot of the average number of PLO to the size of the search space~($|X|=2^{18}$)
\textit{vs.} the average length of the Pareto adaptive walk.
The problem size is $N=18$.\label{fig:nbPOL_length}}
\end{center}
\end{figure}

We performed $10^3$ independent PHC executions for each problem instance.
Fig. \ref{fig:lengthPHC} shows the average length of the Pareto adaptive walks 
for different landscapes according to the set of parameters given in Table \ref{tab:param}. 
% figure
The variation of the average length follows the opposite variation of the number of PLO.
In order to show the link with the number of PLO more clearly, 
Fig. \ref{fig:nbPOL_length} gives the scatter-plot 
of the average Pareto adaptive length \textit{vs.} the logarithm of the average number of PLO.
% figure
The correlation is strong ($r = 0.997$), %$r = 0.996854$
and the regression line equation is:
$\log(y) = a x + b$ , with $a=-1.095$ and $b=12.443$.
% $\log(y) = \exp(-1.09538 x + 12.443)$
For bit-string of length $N=18$,
the average length of the Pareto adaptive walks can then give a precise estimation of the average number of PLO.
When the adaptive length is short, 
the diameter of the basin of attraction associated with a PLO is short.
This means that the distance between PLO decreases.
Moreover, assuming that the volume of this basin is proportional to a power of its diameter,
the number of PLO increases exponentially when the adaptive length decreases.
This corroborates known results from single-objective optimization.
Of course, for larger bit-string length, the coefficients are probably different. 
%Nevertheless, for the same bit-string size, the length of adaptive walks gives a sharp estimator of the number of PLO.

%*************************************************************************************
%\section{Properties \textit{vs.} Multi-modality for Large-size $\rho MNK$-Landscapes}
\section{Properties \textit{vs.} Multi-modality for Large-size Problems}
\label{sec:large}
%*************************************************************************************

In this section, we study the number of PLO for large-size $\rho MNK$-landscapes
using the length of the adaptive walk proposed in the previous section.
First, we analyze this number according to the problem dimension ($N$).
Then, we precise the difficulty, in terms of PLO, with respect to objective space dimension ($M$) and objective correlation ($\rho$).

\begin{figure}[t]
\begin{center}
\begin{tabular}{cc}
\includegraphics[width=0.45\textwidth]{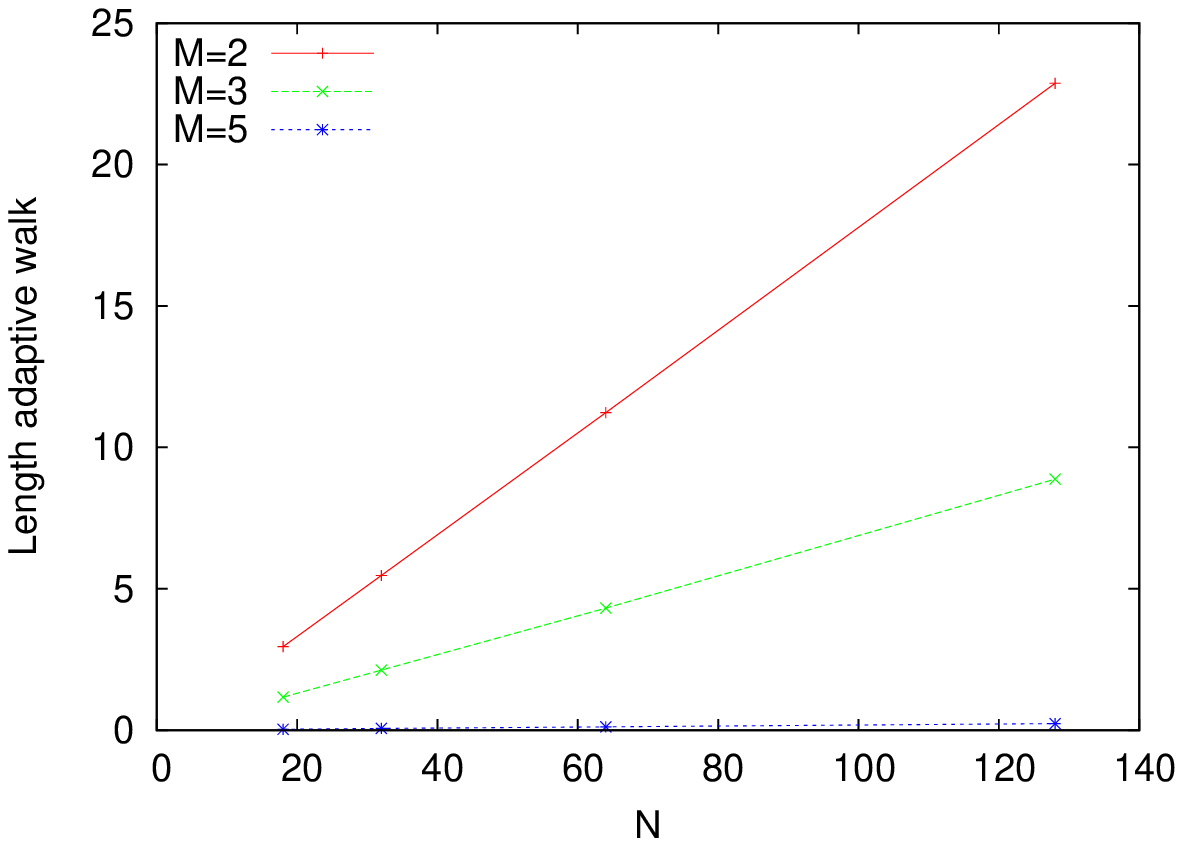} & \includegraphics[width=0.45\textwidth]{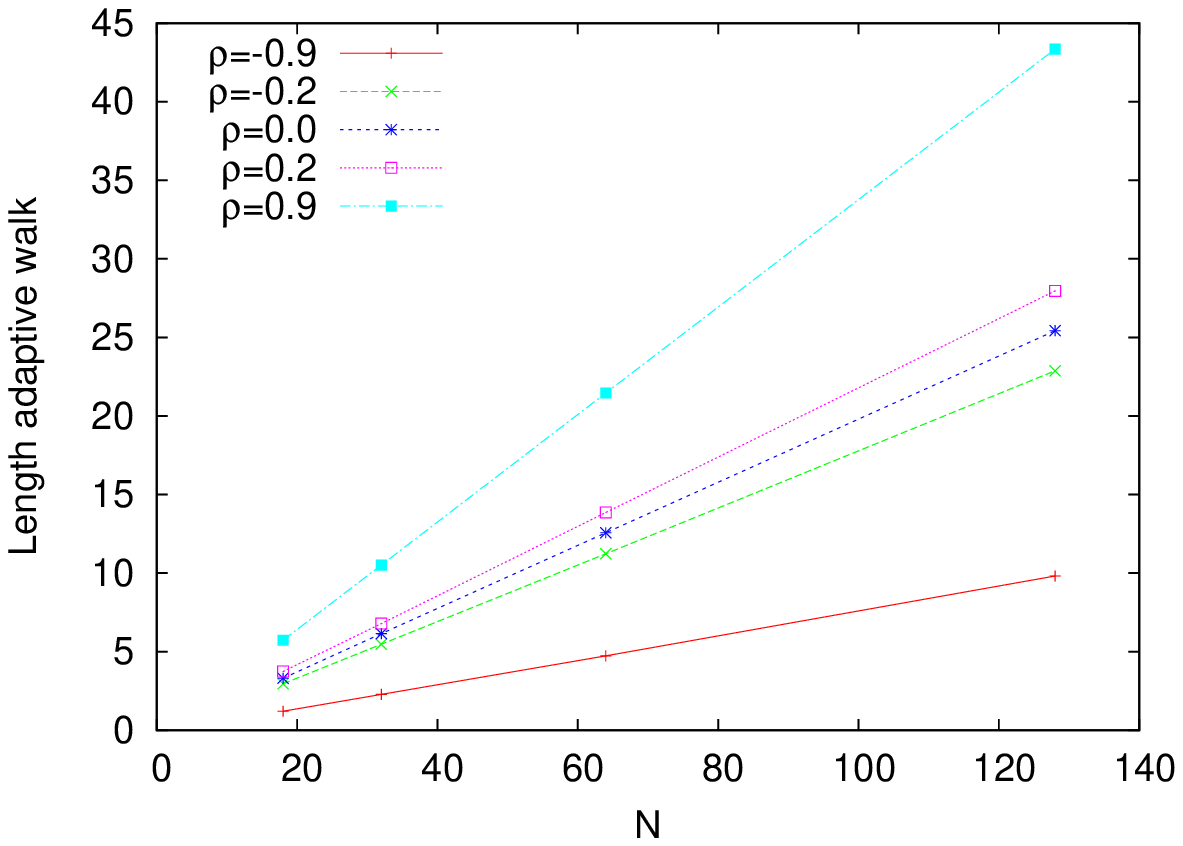} \\
\includegraphics[width=0.45\textwidth]{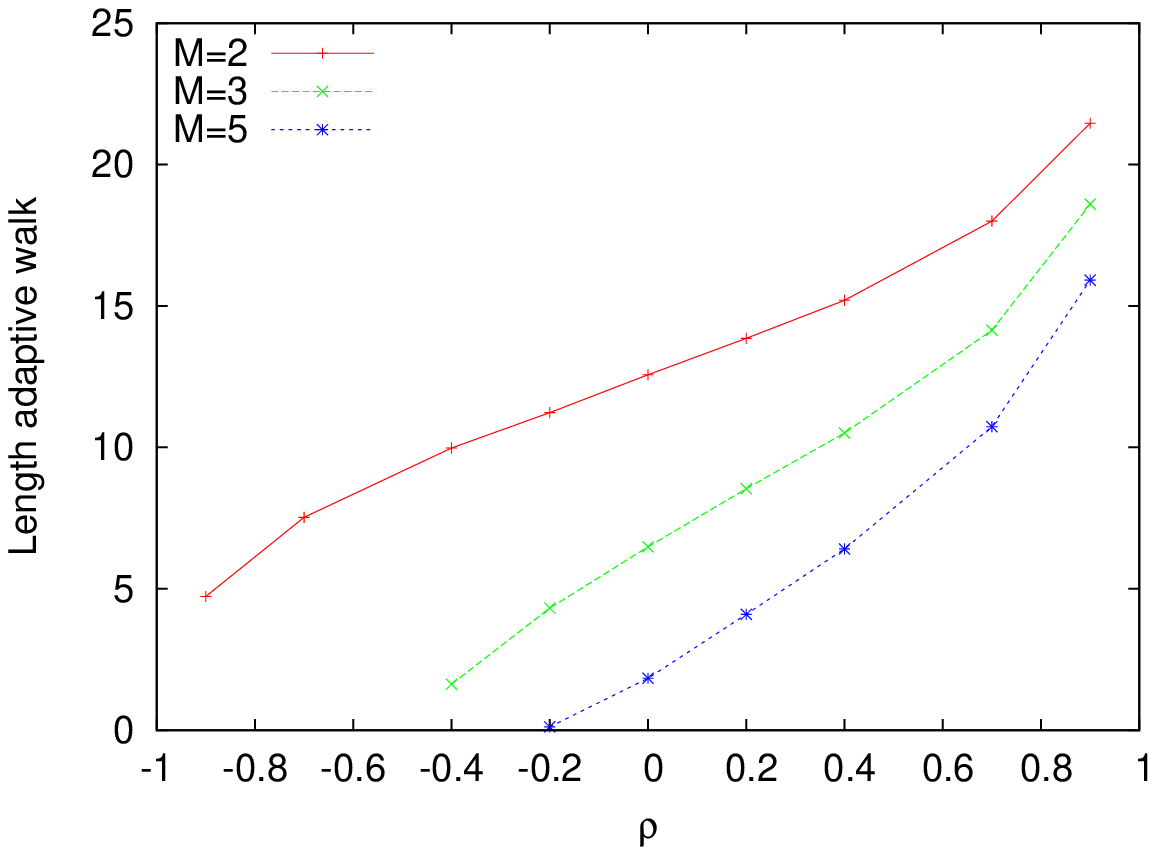} & \includegraphics[width=0.45\textwidth]{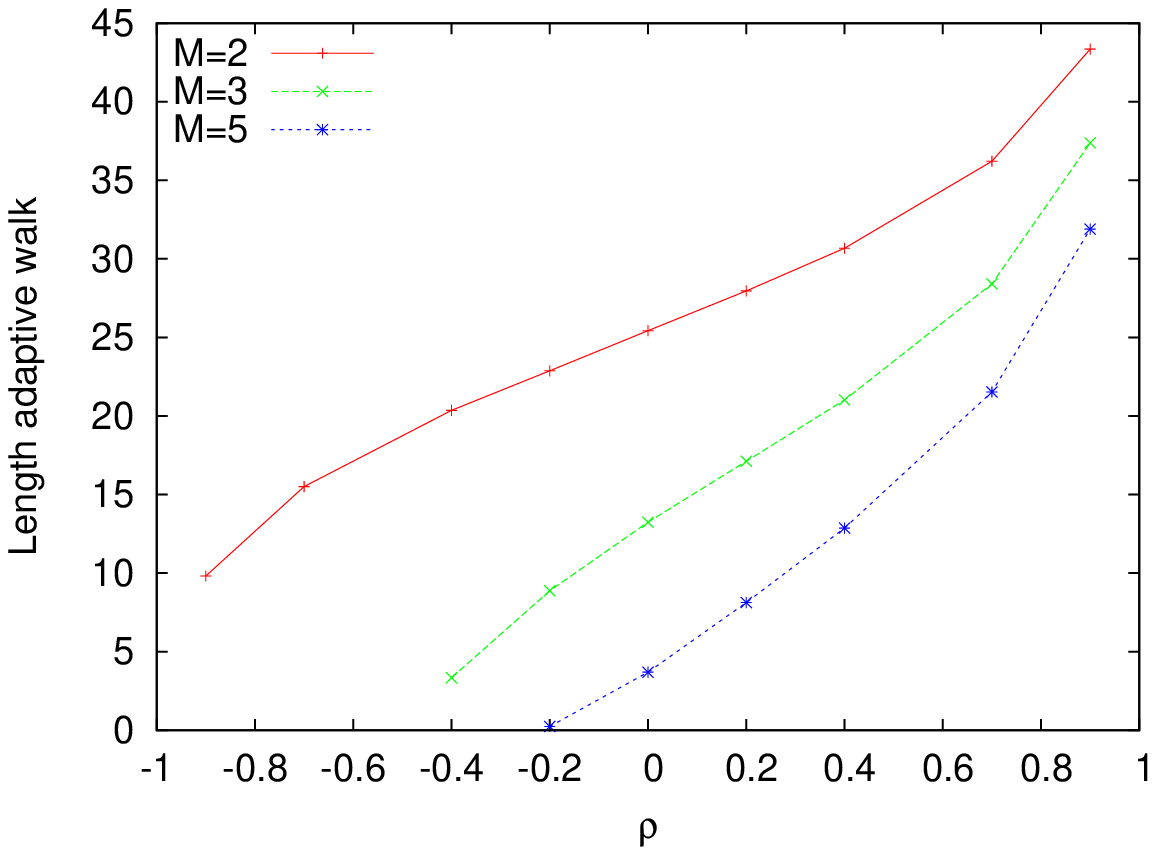} \\
\end{tabular}
\caption{Average length of the Pareto adaptive walk
according to problem size ($N$) for $K=4$ and $\rho=-0.2$ (top-left) and for $K=4$ and $M=2$ (top-right).
Average length of the Pareto adaptive walk according to objective correlation ($\rho$)
for $K=4$ and $N=64$~(bottom-left) and for $K=4$ and $N=128$ (bottom-right). 
\label{fig:lengthPHClarge}}
\end{center}
\end{figure}

We performed $10^3$ independent PHC executions for each problem instance.
Fig. \ref{fig:lengthPHClarge} shows the average length of the Pareto adaptive walks 
for different landscapes according to the set of parameters given in Table \ref{tab:param}. 
Whatever the objective space dimension and correlation, 
the length of the adaptive walks increases linearly with the search space dimension $N$.
According to the results from the previous section, the number of PLO increases exponentially.
We can then reasonably conclude that the size of the Pareto optimal set grows exponentially as well,
to an order of magnitude (Section \ref{sec:est}).
However, the slope of the Pareto adaptive length increase is related to the objective space dimension ($M$) and correlation ($\rho$).
The higher the number of objective functions, the smaller the slope.
As well, the higher the objective correlation, the smaller the slope.

Fig. \ref{fig:lengthPHClarge} (bottom) allows us to give a qualitative comparison for given problem sizes ($N=64$ and $N=128$).
Indeed, let us consider an arbitrary adaptive walk of length $10$.
For $\rho MNK$-landscapes with $N=64$ and $K=4$,
this length corresponds approximately to parameters $(\rho=-0.4,M=2)$, $(\rho=0.3,M=3)$, and $(\rho=0.7,M=5)$ at the same time.
For $N=128$, we have $(\rho=-0.9,M=2)$, $(\rho=-0.1,M=3)$, and $(\rho=0.3,M=5)$.
Still assuming that a problem difficulty is closely related to the number of PLO,
an instance with a small objective space dimension and a negative objective correlation 
can be more difficult to solve than with many correlated objectives. 

% figure

%======================================================================
\section{Discussion}
\label{sec:discs}

This paper gives a fitness landscape analysis for multiobjective combinatorial optimization
based on the local optima of multiobjective $NK$-landscapes with objective correlation.
% Summary
We first focused on small-size problems with a study of the number of local optima by complete enumeration.
Like in single-objective optimization, the number of local optima increases with the degree of non-linearity of the problem (epistasis).
However, the number of objective functions and the objective correlation have a stronger influence.
Futhermore, our results show that the cardinality of the Pareto optimal set clearly increases with the number of local optima.
We proposed a Pareto adaptive walk, associated with a Pareto hill-climber, to estimate the number of local optima for a given problem size.
Next, for large-size instances, the length of such Pareto adaptive walk can give a measure related to the difficulty of a
multiobjective combinatorial optimization problem.
We show that this measure increases exponentially with the problem size.
A problem with a small number of negatively correlated objectives gives the same degree of multi-modality, in terms of Pareto dominance,
than another problem with a high objective space dimension and a positive correlation.

% Future works
A similar analysis would allow to better understand the structure of the landscape for other multiobjective combinatorial optimization problems.
However, an appropriate model to estimate the number of local optima for any problem size still needs to be properly defined.
A possible path is to generalize the approach from~\cite{eremeev2003} for the multiobjective case.
For a more practical purpose,
our results should also be put in relation with the type of the problem under study,
%(permutation, allocation, structural problems\dots)
%in order to give more practical guidelines,
in particular on how to compute or estimate the problem-related measures reported in this paper. 
Moreover, 
%
%In this paper, 
we mainly focused our work on the number of local optima.
The next step is to analyze their distribution by means of a local optima network~\cite{Daolio2010}.
At last, we already know that the number and the distribution of local optima have a strong impact on the performance of multiobjective metaheuristics, but it is not yet clear how they exactly affect the search.
% Open issue
This open issue constitutes one of the main challenge in the %emerging
field of fitness landscape analysis for multiobjective combinatorial optimization.

%======================================================================
\bibliographystyle{splncs}

%\bibliography{rhoMNK,arnaud}

\end{document}